\relax

\documentclass{article}
\usepackage{ijcai19}

\usepackage{amsmath,amsfonts,bm}

\def\eqref#1{equation~\ref{#1}}

\def\1{\bm{1}}

\DeclareMathAlphabet{\mathsfit}{\encodingdefault}{\sfdefault}{m}{sl}
\SetMathAlphabet{\mathsfit}{bold}{\encodingdefault}{\sfdefault}{bx}{n}

\newcommand{\KL}{\mathrm{KL}}

\newcommand{\citet}[1]{\citeauthor{#1}~\shortcite{#1}}
\newcommand{\citep}{\cite}

\usepackage{ifxetex}
\usepackage{subfigure}
\ifxetex
 \usepackage{fontspec}
\else
 \usepackage[T1]{fontenc}
 \usepackage[utf8]{inputenc}
\fi

\usepackage{times}

\usepackage{xspace}
\newcommand*{\eg}{e.g.\@\xspace}
\newcommand*{\ie}{i.e.\@\xspace}

\makeatletter
\newcommand*{\etc}{%
    \@ifnextchar{.}%
        {etc}%
        {etc.\@\xspace}%
}
\makeatother
\usepackage{url}            
\usepackage{booktabs}       
\usepackage{amsfonts}       
\usepackage{nicefrac}       
\usepackage{microtype}      
\usepackage{makecell}       
\usepackage{graphicx}

\usepackage{amsmath,amssymb}
\usepackage{tabularx}
\usepackage{lscape}
\usepackage{mathtools}

\usepackage[capitalise]{cleveref}

\DeclareMathOperator{\Exp}{\mathbb{E}}

\newcommand{\tdot}[3]{\ensuremath{\langle #1, #2, #3 \rangle}}

\DeclareMathOperator{\ELBO}{ELBO}

\newcommand{\Real}{\ensuremath{\mathbb{R}}}

\newcommand{\fs}{\ensuremath{\phi}}

\newcommand{\kg}[1]{{\textsc{#1}}}
\newcommand{\card}[1]{{|#1|}}

\newcommand{\mdl}[1]{{\textsc{#1}}}

\newcommand{\ReP}[1]{\ensuremath{\text{Re}\left(#1\right)}}

\newcommand{\conjt}[1]{\ensuremath{\overline{#1}}}

\newcommand{\KG}{\ensuremath{\mathcal{G}}}

\newcommand{\params}{\ensuremath{\Theta}}
\newcommand{\hparams}{\ensuremath{\Gamma}}
\newcommand{\h}{\ensuremath{\mathbf{h}}}

\usepackage{graphicx}
\usepackage{comment}

\usepackage{xargs}
\usepackage[capitalise]{cleveref}

\newcounter{trCounter}
\newif\iftrvar
\trvartrue
\iftrvar
\newcommand{\tr}[1]{{\small \color{blue} \refstepcounter{trCounter}\textsf{[TR]$_{\arabic{trCounter}}$:{#1}}}}
\else
\newcommand{\tr}[1]{}
\fi

\usepackage{tikz}
\usetikzlibrary{bayesnet}

\usepackage{times}
\usepackage[utf8]{inputenc}
\usepackage[small]{caption}
\usepackage[ruled]{algorithm2e}
\usepackage[noend]{algpseudocode}

\usepackage{autobreak}

\usepackage{color,soul}
\newtheorem{prop}{Proposition}
\usepackage{dsfont}
\usepackage{bm}
\usepackage[inline]{enumitem}
\usepackage{multirow}
\usepackage{amsfonts}

\usepackage{adjustbox}

\definecolor{orange}{rgb}{1,0.5,0}

\title{Neural Variational Inference For Estimating Uncertainty \\ in Knowledge Graph Embeddings}

\author{
Alexander I. Cowen-Rivers$^1$ \and
Pasquale Minervini$^2$\and
Tim Rockt\"{a}schel$^{2}$\and \\
Matko Bo\v{s}njak$^{2}$\and 
Sebastian Riedel$^{2}$\and 
Jun Wang$^{1,2}$
\affiliations
$^1$ MediaGamma\\
$^2$ University College London\\
\emails
alexander.cowen-rivers@mediagamma.com
}

\begin{document}
\maketitle
\begin{abstract}
Recent advances in Neural Variational Inference allowed for a renaissance in latent variable models in a variety of domains involving high-dimensional data.
While traditional variational methods derive an analytical approximation for the intractable distribution over the latent variables, here we construct an inference network conditioned on the symbolic representation of entities and relation types in the Knowledge Graph, to provide the variational distributions.
The new framework results in a highly-scalable method.
Under a Bernoulli sampling framework, we provide an alternative justification for commonly used techniques in large-scale stochastic variational inference, which drastically reduce training time at a cost of an additional approximation to the variational lower bound. 
We introduce two models from this highly scalable probabilistic framework, namely the Latent Information and Latent Fact models, for reasoning over knowledge graph-based representations.
%
Our Latent Information and Latent Fact models improve upon baseline performance under certain conditions. 
%
We use the learnt embedding variance to estimate predictive uncertainty during link prediction, and discuss the quality of these learnt uncertainty estimates. 
%
Our source code and datasets are publicly available online~\footnote{\url{https://github.com/alexanderimanicowenrivers/Neural-Variational-Knowledge-Graphs}}.
\end{abstract}

%
%
\section{Introduction} \label{sec:introduction}
In many fields, including physics and biology, being able to represent \emph{uncertainty} is of crucial importance~\citep{DBLP:journals/nature/Ghahramani15}.
Considering that neural link prediction models for predicting missing links in Knowledge Graphs are used in a variety of decision making tasks~\citep{bean17}, it would be beneficial to assess the predictive uncertainty of a model.
Where a Knowledge Graph is a set of facts between symbols \ie entities. 
However, a significant shortcoming of current neural link prediction models~\citep{timd,DBLP:conf/icml/TrouillonWRGB16} -- and for the vast majority of neural representation learning approaches -- is their inability to express a notion of uncertainty.
Neural link prediction models typically return only point estimates of parameters and predictions~\citep{DBLP:journals/pieee/Nickel0TG16}, and are trained \emph{discriminatively} rather than \emph{generatively}: they aim at predicting one variable of interest conditioned on all the others, rather than accurately representing the relationships between different variables~\citep{DBLP:conf/nips/NgJ01}. 
In a generative probabilistic model, we could leverage the variance in model parameters and predictions for finding which facts to sample during training, in an Active Learning setting~\citep{DBLP:conf/iccv/KapoorGUD07,DBLP:conf/icml/GalIG17}.
Furthermore, Knowledge Graphs can be very large~\citep{DBLP:conf/kdd/0001GHHLMSSZ14}, and often suffer from incompleteness and sparsity~\citep{DBLP:conf/kdd/0001GHHLMSSZ14}: we deal with this through introducing a novel method for including negative sampling in the estimation of the expected lower bound of our probabilistic models. 
\section{Background} \label{sec:background}
In this work, we focus on models for \emph{predicting missing links} in large, multi-relational networks such as \kg{Freebase}, between symbolic items, i.e. nodes. 
In the literature, this problem is referred to as \emph{link prediction}.
We specifically focus on \emph{knowledge graphs}, \ie{}, graph-structured knowledge bases where factual information is stored in the form of relationships between entities.
Link prediction in knowledge graphs is also known as \emph{knowledge base completion}.
We refer to \citep{DBLP:journals/pieee/Nickel0TG16} for a recent survey on approaches to this problem.
A knowledge graph $\KG = \{ (s,r,o) \} \subseteq [N_e] \times [N_r] \times [N_e]$ can be formalised as a set of triples (facts) consisting of a \emph{relation} type $r \in [N_r]$ and two entities $s,o \in [N_e]$, respectively referred to as the \emph{subject} (or \emph{head}) and the \emph{object} (or \emph{tail}) of the triple.
Each knowledge graph triple $(s, r, o)$ encodes a relationship of type $r$ between entities $s$ and $o$. 
A knowledge graph $\KG$ can be represented as an \emph{adjacency tensor} $T \in \{0, 1\}^{|[N_e]| \times |[N_r]| \times |[N_e]|}$, where $T_{s,r,o} = 1$ iff $(s, r, o) \in \KG$, and $T_{s,r,o} = 0$ otherwise.
\emph{Link prediction} in knowledge graphs is often simplified to a \emph{learning to rank} problem, where the objective is to find a score or ranking function $\fs^{\params}_r : [N_e] \times [N_e] \mapsto \Real$ for a relation $r$ that can be used for ranking triples according to the likelihood that the corresponding facts hold true.
\subsection{Neural Link Prediction }
Recently, a specific class of link predictors received a growing interest~\citep{DBLP:journals/pieee/Nickel0TG16}. 
These predictors can be understood as multi-layer neural networks where, given a triple $(s,r,o)$ of symbols, the associated score $\fs^{\params}(s,r,o)$ is given by a neural network architecture encompassing an \emph{encoding layer} and a \emph{scoring layer}. 
In the encoding layer, the subject and object entities $s$ and $o$ are mapped to low-dimensional vector representations (embeddings) $E_s = \h(s) \in \Real^k$ and $E_o = \h(o) \in \Real^k$, produced by an encoder $\h^{\hparams} : [N_e] \to \Real^k$ with parameters $\hparams$. Similarly, relations $r$ are mapped to $R_r = \h(r) \in \Real^k$.
This layer can be pre-trained~\citep{DBLP:conf/acl/VylomovaRCB16} or, more commonly, learnt from data by back-propagating the link prediction error to the encoding layer~\citep{DBLP:journals/pieee/Nickel0TG16,DBLP:conf/icml/TrouillonWRGB16}.
The scoring layer captures the interaction between the entity and relation representations $E_s$, $E_o$ and $R_r$ are scored by a function $\phi^{\params}(E_s,R_r, E_o)$, parametrised by $\params$. Other work encodes the entity-pair in one vector \citep{DBLP:conf/naacl/RiedelYMM13}.
Summarising, the high-level architecture is defined as:
\begin{equation*}
 \begin{aligned}
  E_s, R_r, E_o & = \h^{\hparams}(s), \h^{\hparams}(r), \h^{\hparams}(o) \\
  X_{s,r,o} & \approx \fs(s,r,o) = \fs^{\params}( E_s, R_r, E_o).
 \end{aligned}
\end{equation*}
%
%
Ideally, more likely triples should be associated with higher scores, while less likely triples should be associated with lower scores.
While the literature has produced a multitude of encoding and scoring strategies, for brevity, we overview only a small subset of these. However, we point out that our method makes no further assumptions about the network architecture other than the existence of an encoding layer.   

\paragraph{DistMult.}
\mdl{DistMult}~\citep{yang15:embedding} represents each relation $r$ and entities $s,o$ using parameter vectors $E_s, R_r, E_o \in \Real^{k}$.
For a fact $(s, r, o)$, the model scores the embeddings $(E_s, R_r, E_o)$ using the following scoring function:
\begin{equation*}
\fs^{\params}(E_s, R_r, E_o) = \tdot{R_r}{E_s}{E_o} 
\end{equation*}
\noindent where $\tdot{\cdot}{\cdot}{\cdot}$ denotes the tri-linear dot product.

\paragraph{ComplEx.}

\mdl{ComplEx}~\citep{DBLP:conf/icml/TrouillonWRGB16} is an extension of \mdl{DistMult}~\citep{yang15:embedding} using complex-valued embeddings while retaining the mathematical definition of the dot product.
In this model, the scoring function is defined as:
\begin{equation*}
\fs^{\params}(E_s, R_r, E_o) = \ReP{\tdot{R_r}{E_s}{\conjt{E_o}}},
\end{equation*}
\noindent where $E_s, R_r, E_o \in \mathbb{C}^{k}$ are complex-valued vectors, $\ReP{\cdot}$ denotes the real part of a vector, and $\conjt{E_o}$ denotes the complex conjugate of $E_o$.
%
\section{Generative Models}
In the following, we propose two generative models for knowledge graph embeddings -- the Latent Information Model (LIM) and the Latent Fact Model (LFM).

\begin{figure}
  \centering
  \subfigure[LIM]{\begin{tikzpicture}

  \node[obs]                               (y) {$y$};
 \node[obs]            (t) {$o_i$};
  \node[latent, above=of t, xshift=-2cm] (e) {$E_e$};
  \node[latent, below=of e]  (r) {$E_r$};
  \node[obs, right=of e]            (hr) {$(s_i,r_i)$};

  \edge {e,r,hr} {t} ; %

  \plate {} {(e)} {$e \in [N_e]$} ;
  \plate {} {(r)} {$r \in [N_r]$} ;
  \plate {} {(t)(hr)} {$i \in [n]$} ;


\end{tikzpicture}\label{fig:LIM}}
  \subfigure[LFM]{






\begin{tikzpicture}

  \node[obs]                               (y) {$y$};
  \node[latent, above=of t, xshift=-2cm] (e) {$\mathcal{H}_{h}$};
  \node[obs]            (t) {$o_i$};
  \node[obs, right=of e]            (hr) {$(s_i,r_i)$};

  \edge {e,hr} {t} ; %

  \plate {} {(e)} {$h \in [N_e] \cup [N_r]$} ;
  \plate {} {(t)(hr)} {$i \in [n]$} ;


\end{tikzpicture}\label{fig:LFM}}
\end{figure}
%
%

%
%
%

%
\paragraph{Generative Processes.}
A plate model for the LIM is shown in Figure \ref{fig:LIM}.
%
%
%
Let $\mathcal{D} \subseteq [N_e] \times [N_r] \times [N_e]$ denote a set of triples.
We can define a joint probability distribution over $p(\mathcal{D}, \mathbf{E}, \mathbf{R})$ -- where $\mathbf{E}, \mathbf{R}$ denote all the entity and relation embeddings -- via the following generative model.
\begin{itemize}
    \item For each entity $e \in [N_e]$, and relation $r \in [N_r]$, draw an embedding vector $E_e \sim p(E_e)$ and $R_r \sim p(R_r)$, \eg from a multivariate normal distribution. 
    \item Repeat for each triple $(s, r, o) \in \mathcal{D}:$ \begin{itemize}
        \item Draw a head $s \sim p(s, r)$ and a relation $r \sim p(s, r)$ from the discrete joint distribution $p(s, r)$. The choice of probability distribution $p(s, r)$ has no influence on inference. 
        \item Draw $o \sim \text{Multinomial}(\text{softmax}(X_{s, r, o}))$ with $\text{softmax}(X_{s, r, o}) = \exp(X_{s, r, o}) / \sum_{o'} \exp(X_{s, r, o'})$, where $X_{s,r,o}$ is a model dependent function of $E_s, R_r$ and $E_o$, e.g a function of the model \mdl{ComplEx} $X_{s,r,o} \fs^{\params}(E_s,R_r,E_o)$.
    \end{itemize}
\end{itemize}

\paragraph{Generative Process: LFM}~Fig~\ref{fig:LFM}: A similar generative process to LIM, where we treat the embeddings for the entity and relation embeddings as a single latent variable.   
%
%
%
\subsection{Latent Fact Model} \label{ssec:modelA}
%
%
%
%
%
%
%
The set of latent variables in this model is $\mathbf{H} = \{ \mathbf{E} \cup \mathbf{R} \}$. For the Latent Fact Model (LFM), we assume that the Knowledge Graph was generated according to the following generative model. We place the unit Gaussian prior $p^{\theta}(\mathbf{H}) =\mathcal{G}(0,\mathcal{I})$ on $\mathbf{H}$.
The joint probability of the variables $p^{\theta}(\mathcal{D}, \mathbf{H})$ is defined as follows:
\begin{equation*}
 p^{\theta}(\mathcal{D}, \mathbf{H}) = \prod_{h \in [N_e] \cup [N_r]} p^{\theta}(\mathbf{H}_{h}) \prod_{(X_{s,r,o}) \in \mathcal{D}}  p^{\theta}(X_{s,r,o} \mid \mathbf{H}_{h})
\end{equation*}
The marginal distribution over $\mathcal{D}$ is then bounded as follows, with respect to our variational distribution $q$:
\begin{prop}
As a consequence, the log-marginal likelihood of the data, under the Latent Fact Model, is bounded by:
\begin{equation}\label{eq:elboA}
\begin{aligned}
  & \log p^{\theta}(\mathcal{D}) \geq \\
  & \quad \Exp_{\mathbf{H} \sim q^{\phi}} \left[ \log p^{\theta}(\mathcal{D} \mid \mathbf{H}) \right] - \KL[{ q^{\phi}(\mathbf{H}) } \mid\mid{ p^{\theta}(\mathbf{H}) }] &
\end{aligned}
\end{equation}
\end{prop}
%
%

\paragraph{Assumptions:} LFM model assumes each \emph{fact} of is a randomly generated variable, as well as a mean field variational distribution and that each training example is independently distributed. 
\subsubsection{Optimising LFM's ELBO}\label{optm:elbo}
Note that this is an enormous sum over $\card{\mathcal{D}}$ elements, which can be approximated via Importance Sampling, or Bernoulli Sampling~\citep{DBLP:conf/aistats/BotevZB17}.
\begin{equation*}
\begin{aligned}
& \ELBO = \sum_{(X_{s,r,o}) \in \mathcal{D}} \Exp_{\mathbf{E},  \mathbf{R} \sim q^\phi} \left[ \log p^{\theta}(X_{s,r,o} \mid \mathbf{H}) \right] \\
& \qquad - \KL[{ q^{\phi}(\mathbf{H}) } \mid\mid{ p^{\theta}(\mathbf{H}) }] \\
& \quad = \sum_{(X_{s,r,o}) \in \mathcal{D}^{+}} \Exp_{\mathbf{H}  \sim q^\phi} \left[ \log p^{\theta}(X_{s,r,o} \mid \mathbf{H}) \right] \\
& \qquad + \sum_{(X_{s,r,o}) \in \mathcal{D}^{-}} \Exp_{\mathbf{H}  \sim q^\phi} \left[ \log p^{\theta}(X_{s,r,o} \mid \mathbf{H}) \right] \\
& \qquad - \KL[{ q^{\phi}(\mathbf{H}) } \mid\mid{ p^{\theta}(\mathbf{H}) }]\\
\end{aligned}
\end{equation*}
By using Bernoulli Sampling, $\ELBO$ can be approximated by defining a probability distribution of sampling from $\mathcal{D}^{+}$ and $\mathcal{D}^{-}$  -- similarly to Bayesian Personalised Ranking~\citep{DBLP:conf/uai/RendleFGS09}, we sample one negative triple for each positive one --- we use a constant probability for each element depending on whether it is in the positive or negative set. 
%
\begin{prop}
The Latent Fact models $\ELBO$ can be estimated similarly using a constant probability for positive or negative samples. We end up with the following estimate:
\begin{equation*} 
\begin{aligned}
  & \ELBO \approx \sum_{(X_{s,r,o}) \in \mathcal{D}^{+}}  \frac{s_{s,r,o}}{b^{+}}  \ \ \Exp_{\mathbf{H}  \sim q^\phi} \left[ \log p^{\theta}(X_{s,r,o} \mid \mathbf{H}) \right] \\
  & \quad + \sum_{(X_{s,r,o}) \in \mathcal{D}^{-}} \frac{s_{s,r,o}}{b^{-}} \ \ \Exp_{\mathbf{H}  \sim q^\phi} \left[ \log p^{\theta}(X_{s,r,o} \mid \mathbf{H}) \right] \\
  & \quad - \KL[{ q^{\phi}(\mathbf{H}) } \mid\mid{ p^{\theta}(\mathbf{H}) }]\ \  \\
\end{aligned}
\end{equation*}
\end{prop}
\noindent where $p^{\theta}(s_{s,r,o} = 1) = b_{s,r,o}$ can be defined as the probability that for the coefficient $s_{s,r,o}$ each positive or negative fact ${s,r,o}$ is equal to one (i.e is included in the ELBO summation). The exact ELBO can be recovered from setting $b_{{s,r,o}}=1.0$ for all ${s,r,o}$.
\noindent where $b^{+} = \card{\mathcal{D}^{+}} / \card{\mathcal{D}^{+}}$ and $b^{-} = \card{\mathcal{D}^{+}} / \card{\mathcal{D}^{-}}$.
\subsection{Latent Information Model} \label{ssec:modelB}
%
%
In Figure~\ref{fig:LIM}'s graphical model, we assume that the Knowledge Graph was generated according to the following generative model. The set of latent entity variables in this model is $\mathbf{E} = \{ E_{e} \mid e \in [N_e] \}$ and the set of latent relation variables $\mathbf{R} = \ \{ R_{r} \mid r \in [N_r] \}$.  We place the following unit Gaussian priors $p^{\theta}(E) =\mathcal{G}(0,\mathcal{I})$ and $p^{\theta}(R) =\mathcal{G}(0,\mathcal{I})$ on $E$ and $R$, respectively. The joint probability of the variables $p^{\theta}(\mathcal{D}, \mathbf{E},  \mathbf{R})$ is defined as follows:
%



%
%
%
%
%
\begin{equation}
\begin{aligned}
  & p^{\theta}(\mathcal{D}, \mathbf{E},  \mathbf{R}) \\ &= \prod_{e \in [N_e]} p^{\theta}(E_{e}) \prod_{r \in [N_r]} p^{\theta}(R_{r}) \prod_{(X_{s,r,o}) \in \mathcal{D}} p^{\theta}(X_{s,r,o} \mid \mathbf{E},  \mathbf{R})
\end{aligned}
\end{equation}
%
%
\begin{prop}
    The log-marginal likelihood of the data, under the Latent Information Model, is the following:

\begin{equation}
\begin{aligned}
  \log p^{\theta}(\mathcal{D}) \geq & \Exp_{\mathbf{E}, \mathbf{R}  \sim q^\phi} \left[ \log p^{\theta}(\mathcal{D} \mid \mathbf{E},  \mathbf{R}) \right] \\ & - \KL[{ q^{\phi}(\mathbf{E}) } \mid\mid{ p^{\theta}(\mathbf{E}) }] - \KL[{ q^{\phi}(\mathbf{R}) } \mid\mid{ p^{\theta}(\mathbf{R}) }]
\end{aligned}
\end{equation}
\end{prop}
%
\paragraph{Assumptions:} LIM makes the same assumptions as LFM, with the additional assumption that the entities and relations are separate latent variables. 
%
\subsubsection{Optimising LIM's ELBO}

Similarly to Section~\ref{optm:elbo}, by using Bernoulli Sampling the $\ELBO$ can be approximated by using a constant probability for positive or negative samples, we end up with the following estimate:
\begin{prop}
        The Latent Information Models $\ELBO$ can be estimated similarly using a constant probability for positive or negative samples. We end up with the following estimate:

\begin{equation} 
\begin{aligned}
  & \ELBO \approx \\ & ( \sum_{(X_{s,r,o}) \in \mathcal{D}^{+}}  \frac{s_{s,r,o}}{b^{+}} \ \ \Exp_{\mathbf{E},  \mathbf{R} \sim q^\phi} \left[ \log p^{\theta}(X_{s,r,o} \mid \mathbf{E}, \mathbf{R}) \right]) \\
  & + ( \sum_{(X_{s,r,o}) \in \mathcal{D}^{-}} \frac{s_{s,r,o}}{b^{-}} \ \ \Exp_{\mathbf{E},  \mathbf{R} \sim q^\phi} \left[ \log p^{\theta}(X_{s,r,o} \mid \mathbf{E}, \mathbf{R}) \right]) \\
  & - \KL[{ q^{\phi}(\mathbf{E}) } \mid\mid{ p^{\theta}(\mathbf{E}) }]\ 
 -  \KL[{ q^{\phi}(\mathbf{R}) } \mid\mid{ p^{\theta}(\mathbf{R}) }]\ 
\end{aligned}
\end{equation}
\end{prop}
\noindent where $b^{+} = \card{\mathcal{D}^{+}} / \card{\mathcal{D}^{+}}$ and $b^{-} = \card{\mathcal{D}^{+}} / \card{\mathcal{D}^{-}}$.


%

%
%
\section{Related Work}

Variational Deep Learning has seen great success in areas such as parametric/non-parametric document modelling~\citep{2017arXiv170600359M,miao_yu_blunsom_2016} and image generation \citep{Kingma2013}.  Stochastic variational inference has been used to learn probability distributions over model weights ~\citep{2015arXiv150505424B}, which the authors named "Bayes By Backprop". These models have proven powerful enough to train deep belief networks ~\citep{DBLP:journals/corr/VilnisM14}, by improving upon the stochastic variational Bayes estimator ~\citep{Kingma2013}, using general variance reduction techniques.

Previous work has also researched word embeddings within a Bayesian framework ~\citep{zhang_salwen_glass_gliozzo_2014,DBLP:journals/corr/VilnisM14}, as well as researched graph embeddings in a Bayesian framework ~\citep{He:2015:LRK:2806416.2806502}. However, these methods are expensive to train due to the evaluation of complex tensor inversions. Recent work by ~\citep{DBLP:journals/corr/Barkan16,2017arXiv171111027B} show that it is possible to train word embeddings through a variational Bayes ~\citep{bishop_2006} framework. 

KG2E ~\citep{He:2015:LRK:2806416.2806502} proposed a probabilistic embedding method for modelling the uncertainties in KGs. However, this was not a generative model. \citep{Xiao2016TransGAF} argued theirs was the first generative model for knowledge graph embeddings. However, their work is empirically worse than a few of the generative models built under our proposed framework, and their method is restricted to a Gaussian distribution prior. In contrast, we can use any prior that permits a re-parameterisation trick --- such as a Normal~\citep{DBLP:journals/corr/KingmaW13} or von-Mises distribution~\citep{davidson2018hyperspherical}.

Later, \citep{vgae} proposed a generative model for graph embeddings. However, their method lacks scalability as it requires the use of the full adjacency tensor of the graph as input. Moreover, our work differs in that we create a framework for many variational generative models over multi-relational data, rather than just a single generative model over uni-relational data \citep{vgae,grover2018graphite}. In a different task of graph generation, similar models have been used on graph inputs, such as variational auto-encoders, to generate full graph structures, such as molecules \citep{simonovsky2018graphvae,liu2018constrained,de2018molgan}. ~\citep{salehi2018probabilistic} recently purposed a probabilistic knowledge graph model, this is then used to learn regularisation weights using EM, whereas we want to focus on studying the learnt predictive uncertainty and not focus on learning a regularisation weight. 
Recent work by \citep{Chen2018VariationalKG} constructed a variational path ranking algorithm, a graph feature model. This work differs from ours for two reasons. Firstly, it does not produce a generative model for knowledge graph embeddings. Secondly, their work is a graph feature model, with the constraint of at most one relation per entity pair, whereas our model is a latent feature model with a theoretical unconstrained limit on the number of existing relationships between a given pair of entities. 
\section{Experiments}\label{results}

\paragraph{Experimental Setup}
We run each link prediction experiment over 500 epochs and validate every 50 epochs. Each KB dataset is separated into 80 \% training facts, 10\% development facts, and 10\% test facts. 
\begin{table}
\centering
\resizebox{\columnwidth}{!}{%

\begin{tabular}{|c|c|ccccc}
\hline
Dataset          & {Scoring Function}                                                            & \multicolumn{2}{c|}{MR}                          & \multicolumn{3}{c|}{Hits @}                                         \\ \cline{3-7} 
                                  &                                                                                   & \multicolumn{1}{c|}{Filter} & \multicolumn{1}{c|}{Raw} & \multicolumn{1}{c|}{1} & \multicolumn{1}{c|}{3} & \multicolumn{1}{c|}{10}    \\ \hline
WN18                                                & V DistMult (LIM)                                                             & 786      & 798                      & 0.671                  & 0.931                  & \multicolumn{1}{c|}{0.947}        \\ & DistMult                                                             & \multicolumn{1}{r}{813}          & 827                      & 0.754                  & 0.911                  & \multicolumn{1}{l|}{0.939}          \\
                                                                              & V ComplEx (LIM)                                                              & \textbf{753}        & \textbf{765}                     & {0.934}                 &{\textbf{0.945}}         & \multicolumn{1}{c|}{\textbf{0.952}}      \\ 
                                                                              & ComplEx*  & \multicolumn{1}{r}{--}            & --                        & \textbf{0.939}         & 0.944         & \multicolumn{1}{l|}{0.947} \\
                                                        
                                                     \hline
WN18RR   & V DistMult (LIM)                                                             & 6095                             & \textbf{6109}                     & 0.357                  & 0.423                 & \multicolumn{1}{c|}{0.440}          \\ & DistMult                                                             & 8595                             & 8595                     & 0.367                  & 0.390                  & \multicolumn{1}{l|}{0.412}          \\
                                                                              & V ComplEx (LFM)                                                              & 6500                             & 6514                     & 0.385         & 0.446         & \multicolumn{1}{c|}{0.489}          \\
                                                                              & ComplEx**  & \textbf{5261}                    & --                        & \textbf{0.41}          & \textbf{0.46}          & \multicolumn{1}{l|}{\textbf{0.51}}  \\
                                                                                \hline
\end{tabular}
}
\caption{Filtered and Mean Rank (MR) for the models tested on the WN18, WN18RR datasets. Hits@m metrics are filtered. Scoring functions with a "V" are results we reported under our variational framework LIM/LFM vs reported baseline results.} 
\label{ch7:linearwarm-up}
\end{table} 
\paragraph{Results}
Table~\ref{ch7:linearwarm-up} shows definite improvements on WN18 for Variational ComplEx compared with the initially published x. We believe this is due to the well-balanced model regularisation induced by the zero mean unit variance Gaussian prior. Table~\ref{ch7:linearwarm-up} also shows that the variational framework is outperformed by existing non-generative models, highlighting that the generative model may be better suited at identifying and predicting symmetric relationships. WordNet18~\citep{NIPS2013_5071} (WN18) is a large lexical database of English. WN18RR is a subset with only asymmetric relations. 
%
%
We now compare our model to the previous state-of-the-art multi-relational generative model TransG ~\citep{Xiao2016TransGAF}, as well as to a previously published probabilistic embedding method KG2E (similarly represents each embedding with a multivariate Gaussian distribution)  ~\citep{He:2015:LRK:2806416.2806502} on the WN18 dataset.
\begin{table}
\centering
\resizebox{\columnwidth}{!}{%
\begin{tabular}{|c|lcll|}
\hline
{Dataset} & \multicolumn{1}{l|}{{Scoring Function}}              & \multicolumn{2}{c|}{MR}       & \multicolumn{1}{l|}{Filtered} \\ \cline{3-4} 
                         & \multicolumn{1}{l|}{}                                    & Raw   & \multicolumn{1}{l|}{Filter} & \multicolumn{1}{l|}{Hits@ 10}        \\ \hline
{WN18}    & KG2E  \citep{He:2015:LRK:2806416.2806502} & \textbf{362}   & \textbf{345}                                                     & 0.932        \\
                         & TransG (Generative) \citep{Xiao2016TransGAF}  & 345   & 357                                                           & 0.949         \\
                         & Variational ComplEx (LIM) & 753   & 765                                                    & \textbf{0.952}     \\ \hline
\end{tabular}
}
\caption{Latent Information Model vs. Existing Generative Models}
\label{analysis:vsTransG}
\end{table}
Table~\ref{analysis:vsTransG} makes clear the improvements in the performance of the previous state-of-the-art generative multi-relational knowledge graph model. 
\begin{figure}[t]
\centering
         \includegraphics[width=\columnwidth]{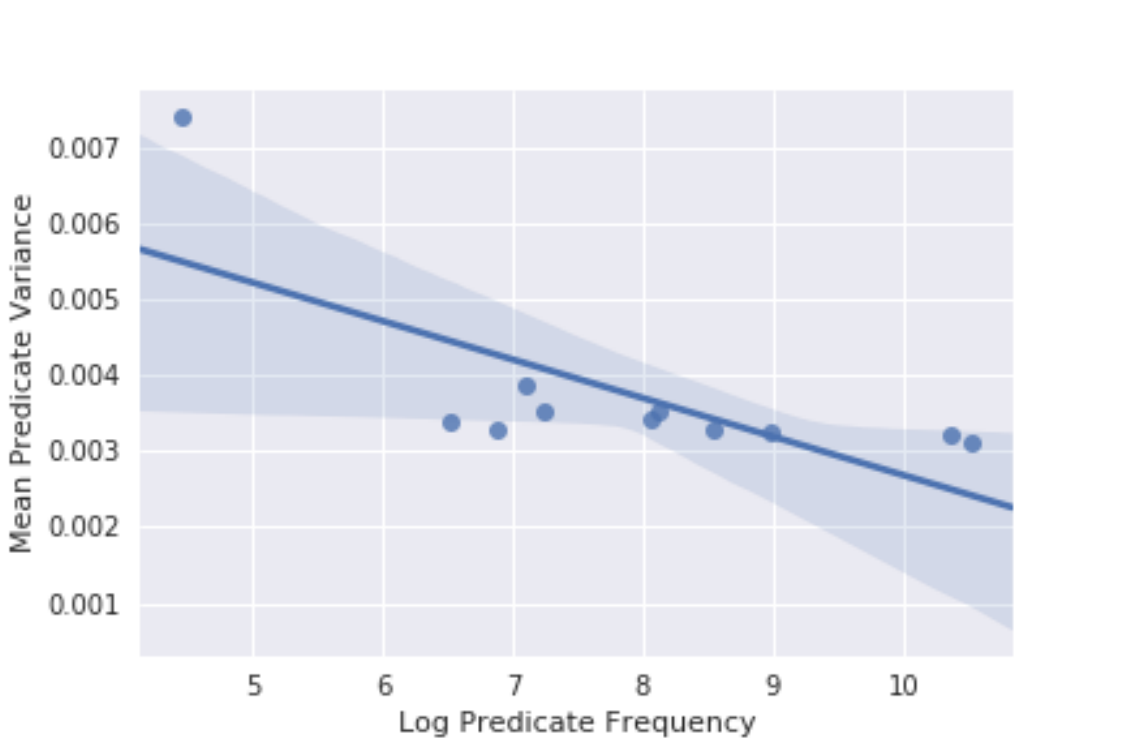}
         \includegraphics[width=\columnwidth]{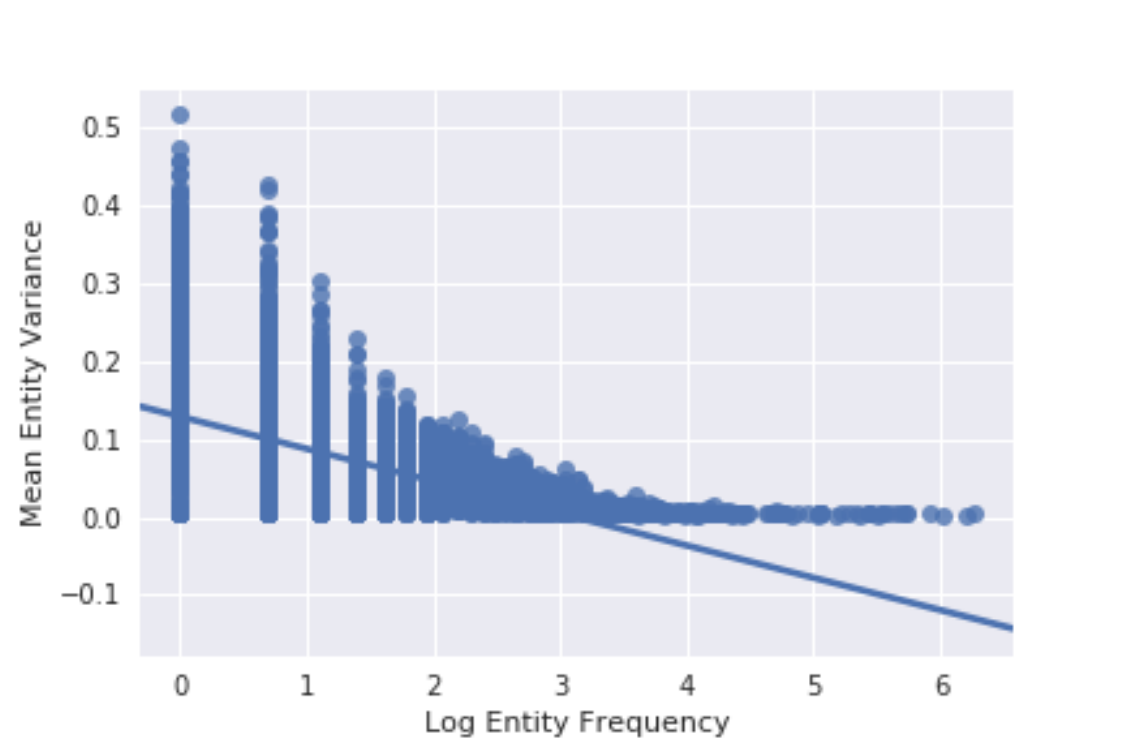}
    \caption{Mean Variance vs. log frequency. Top: WN18RR Predicate Matrix. Bottom: WN18RR Entity Matrix.}
    \label{fig:freqvaranalys}
\end{figure}

\paragraph{Uncertainty Analysis}\label{analysis:embedding_anaysis}
These results hint at the possibility that the slightly stronger results of WN18 are due to covariances in our variational framework able to capture information about symbol frequencies.
We verify this by plotting the mean value of covariance matrices, as a function of the entity or predicate frequencies (Figure~\ref{fig:freqvaranalys}). The plots confirm our hypothesis: covariances for the variational Latent Information Model grows with the frequency, and hence the LIM would put a preference on predicting relationships between less frequent symbols in the knowledge graph. This also suggests that covariances from the generative framework can capture accurate information about the generality of symbolic representations.
%
%
%
%
%
%
%
Motivated by the desiring to reduce predictive uncertainty, we explore two methods for confidence estimation by; taking the magnitude of the prediction as confidence, attempting to measuring the models' predictive uncertainty (achieved through forward sampling). This experiment was carried out using the LIM on Nations dataset with, Variational DistMult. 
%
\begin{figure}[t]
    \centering
    \includegraphics[width=\columnwidth]{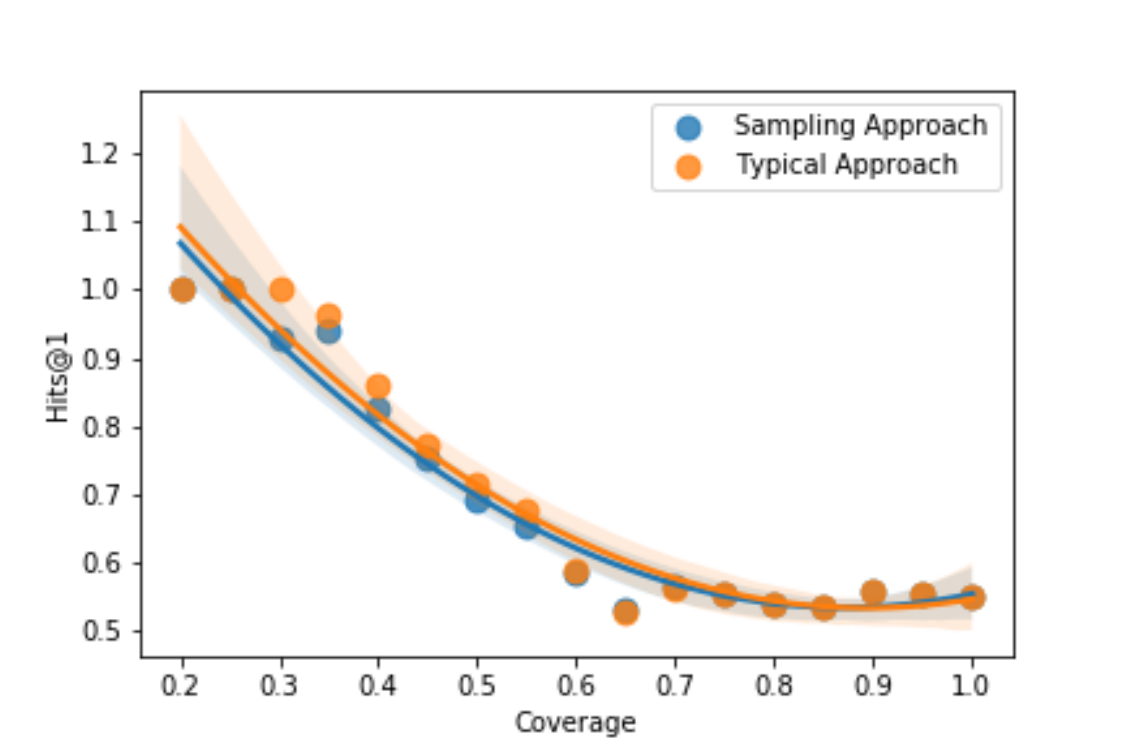}
    \caption{Precision - Coverage Relationship. For the first confidence estimation method, we interpret the magnitude of the prediction as confidence. We search over 1,000 coverage values between (0,1]. At each coverage value, we implement a threshold in which predictions outside this confidence range are discarded. We then plot these and fit a regression line of order two, to estimate the trend.}
    \label{fig:confestim}
\end{figure}
%
Based on Fig~\ref{fig:confestim}, we can see a general trend of increased precision with a decrease in coverage, exactly what we would desire from a model to estimate its confidence in a prediction. Unfortunately, utilising the uncertainty on the latent embeddings through sampling does not result in improved uncertainty estimates over using the magnitude of likelihood estimate as the confidence, which leaves further room for research into how best to utilise these learnt uncertainty estimates. 

\paragraph{Visualised Variational Embedding Distributions}

\begin{figure}[t]
    \centering
    \includegraphics[width=\columnwidth]{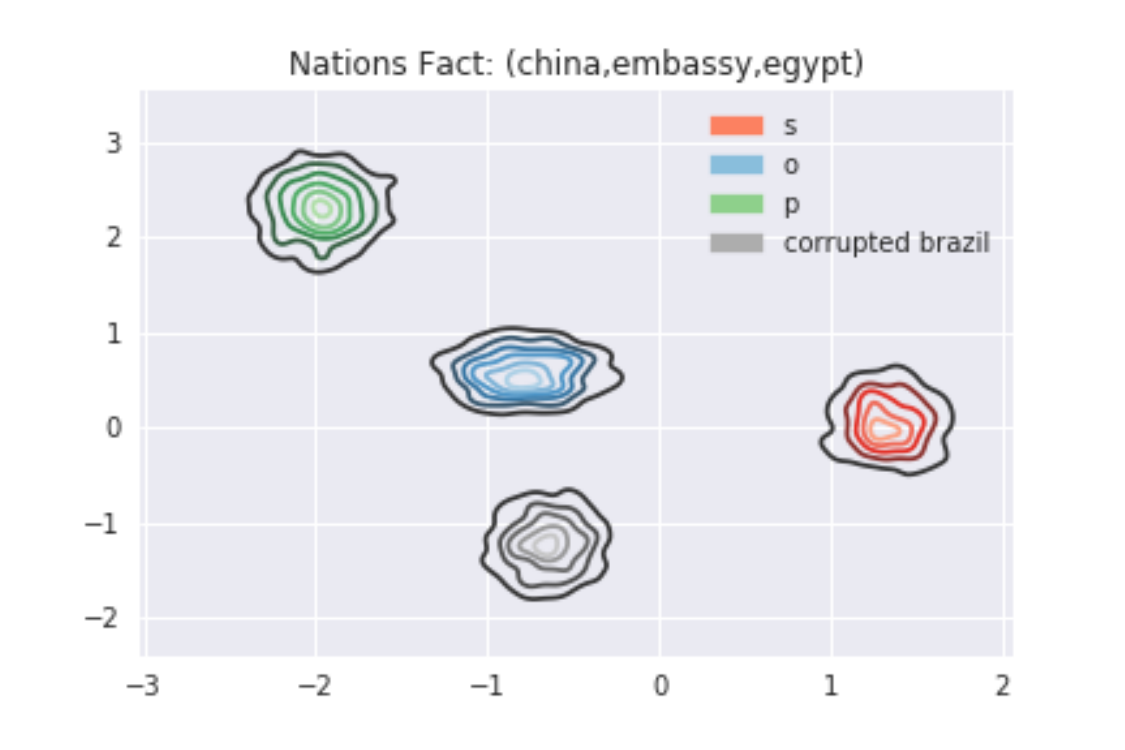}
    \includegraphics[width=\columnwidth]{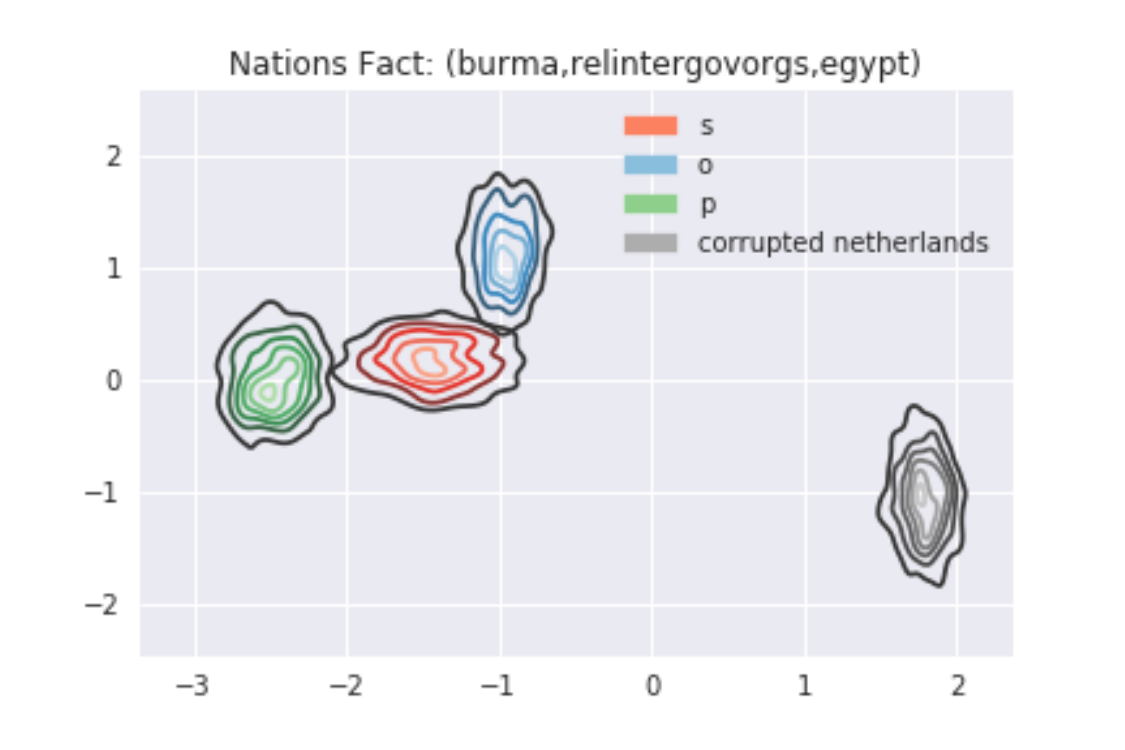}
    \caption{True positives. Each image visualises a facts subject (red), object (blue) and relation (green) embedding, to show similarity, as well as a randomly sampled corrupted embedding to show dissimilarity. Top: $China\Rightarrow^\text{Embassy}Egypt$. Bottom: $Burma\Rightarrow^\text{Intergoveremental Org}Egypt$}
    \label{fig:finalembdstrue}
\end{figure}

We project the high dimensional mean embedding vectors to two dimensions using Principal Component Analysis, to project the variance embedding vectors down to two dimensions using Non-negative Matrix Factorisation. Once we have the parameters for a bivariate normal distribution, we then sample from the bivariate normal distribution 1,000 times and then plot a bi-variate kernel density estimate of these samples. By visualising these two-dimensional samples, we can conceive the space in which the entity or relation occupies. We complete this process for the subject, object, relation, and a randomly sampled corrupted entity (under LCWA) to produce a visualisation of a fact, as shown in Figure~\ref{fig:finalembdstrue}. 
Figure~\ref{fig:finalembdstrue} displays two true positives from test time predictions.  The plots show that the variational framework can learn high dimensional representations which when projected onto lower (more interpretable) dimensions, the distribution over embeddings are shaped to occupy areas at which facts lie.  
\section{Conclusion}
We argue there is a lack of methods for quantifying predictive uncertainty in a knowledge graph embedding representation, which can only be utilised using probabilistic modelling, as well as a lack of expressiveness under fixed-point representations. 
We introduce a framework for creating a family of highly scalable probabilistic models for knowledge graph representation
The framework improves model performance under certain conditions, while reducing the parameter search by one hyper-parameter, as the unit Gaussian prior is self-regularising. Overall, we believe this work will enable knowledge graph researchers to work towards the goal of creating models better able to express their predictive uncertainty.

\section*{Acknowledgments}
We want to thank all members of the UCL NLP for useful discussions, and facilities provided by MediaGamma Ltd. 
\bibliographystyle{named}
{\small \bibliography{ijcai19}}

\end{document}